\documentclass[12pt,twoside]{article}
\usepackage{graphics,color}      % usual driver
\usepackage{verbatim}
\usepackage{amsthm}
\usepackage{html}
\usepackage{hyperref}
\usepackage{float}
\usepackage{fancyhdr}
\usepackage{textcomp}
\usepackage[uline]{hhtensor}
\usepackage{mathtools}
\usepackage[labeled,resetlabels]{multibib}
\usepackage{amsmath}    % need for subequations
\usepackage{psfig} %capability to place postscript drawings from xfig

\hypersetup{colorlinks,urlcolor=blue}

\theoremstyle{definition}
\newtheorem{dfn}{Definition}

\theoremstyle{remark}
\newtheorem*{rmk}{Remark}
\theoremstyle{remark}

\catcode`,\active

\catcode`\,12

\makeatletter
\DeclareRobustCommand{\bardelta}{\mathrel{\mathpalette\bar@delta\relax}}
\newcommand{\bar@delta}[2]{%
  \hspace{1mm} \mathstrut\ooalign{\hidewidth$\m@th#1^{-}$\hidewidth\cr$\m@th#1\delta$\cr}%
}
\makeatother

\begin{document}

\pagestyle{plain}
\begin{center}\begin{Large}Policies for constraining the behaviour of coalitions of agents in the context of algebraic information theory\end{Large}\end{center}% see notes 10 page 14 %
\begin{center}Chris Goddard \\
\today \end{center}
\begin{center} \textbf{Abstract} \end{center}
\begin{center} \small{\emph{This article takes an oblique sidestep from two previous papers, wherein an approach to reformulation of game theory in terms of information theory, topology, as well as a few other notions was indicated.  In this document a description is provided as to how one might determine an approach for an agent to choose a policy concerning which actions to take in a game that constrains behaviour of subsidiary agents. It is then demonstrated how these results in algebraic information theory, together with previous investigations in geometric and topological information theory, can be unified into a single cohesive framework.}} \end{center}

\section{Intuition}

%\begin{quote}
%$\kappa\alpha\iota$ $\epsilon\alpha\nu$ $\omicron\iota\kappa\iota\alpha$ $\epsilon\phi$ $\epsilon\alpha\upsilon\tau\eta\nu$
%$\mu\epsilon\rho\iota\sigma\theta\eta$, $\omicron\upsilon$ $\delta\upsilon\nu\eta\sigma\epsilon\tau\alpha\iota$ $\eta$ 
%$\omicron\iota\kappa\iota\alpha$ $\epsilon\kappa\epsilon\iota\nu\eta$ $\sigma\tau\eta\nu\alpha\iota$.
%
%If a house is divided against itself, that house cannot stand.
%% 
%
%%??? ??? ????? ??Õ ?????? ???????, ?? ????????? ? ????? ?????? ??????.
%
%\begin{flushright}\emph{Mark 3:25}\end{flushright}
%\end{quote}

\subsection{On the centrality of regulation in maintaining economic competitiveness}

Consider the real world situation where agents compete and/or cooperate in order to maximise their share of scarce resources.  Their actions are constrained by laws, the formulation and implementation of which by regulators is driven by policies.

In particular, consider the role that regulatory agencies fulfil in preventing overfishing, or in ensuring that financial integrity in maintained in a financial market, thereby preventing anarchy and corruption (and thereby waste of economic value, at the cost of competitiveness of the society in which the regulatory agency is situated).  Indeed, effective regulation of play in an economic game goes hand in hand with the ability of a society to compete with other societies effectively, for the means of economic production - be that labour or otherwise.

To drive home this point, let us expand upon the example of the tragedy of the commons.  This apocryphal example is of a village with a single green, and various townsfolk with their own separate flocks of sheep.  Each townsperson has the ability to allow their sheep to graze a small patch of ground on the village common.  Most will not intend to deplete the resource, but eventually they will realise that it will be depleted by others if they don't deplete it first, so they race to use it all.  So the grass is depleted, the resource is gone, and the flocks of sheep have a much poorer diet.  This evidently causes all players to lose in an iterated game.  It also 'causes the town to lose' as other towns with policies regarding village greens that are not so lax will attract players from the original town (as said towns will have better land for grazing sheep).  Indeed, a regulator (eg the town) can issue a bounded number of permits for use of the land for a particular amount of time, and at a certain level of depletion, which solves this problem.

Some further intuition may be built by considering the role that regulatory entities fulfil in preventing agents from 'gaming' markets, so that the tragedy of the commons can be avoided.  Inefficient markets allow agents to take actions that transfer costs to other players.  Regulatory entities provide forcing functions that ideally help players wear the true costs of their activity in the game.  Of course, generally speaking, regulators or adjudicators provide some form of influence on the game that acts as a forcing function on payoff functions for particular actions at particular times.

More formally, we would like to study how regulators can act in a way such that their \emph{intervention} is optimal, i.e. such that it prevents and/or addresses market distortions that would arise as emergent behaviours and/or exploits (violations of the efficient market hypothesis) as a consequence of the basic structure of an economic game.  In a way this can be seen as congruent to establishing an ethical framework, as ethics can be seen as concerning itself in terms of how one should regulate behaviour with regards to competition for finite resources in some form of game.

A key goal of this paper will be to construct a framework for the formal consideration of issues such as these.

In particular, within the document, a model will be introduced, within which a system agent regulates the playing of a game, which may or may not be iterated, by coalitions of sub-agents who may compete, coordinate with each other and/or cooperate.  The system agent may constrain the actions of the sub-agents using \emph{laws} in accordance with \emph{policies}.

Throughout said discussion, it will also be indicated how these ideas can be generalised, abstracted, and extended - within the broader and richer context of a holistic vantage regarding first-order considerations in Algebra, Geometry, and Topology.

\subsection{Definitions, agents, constraints}

Now, bear in mind the situation that governed said preceding papers on topological information theory: we have a game (i.e., a decision problem with some form of payoff function, that may or may not be iterated, and which generically involves competition, coordination, and/or cooperation between multiple entities over said payoff), and we have agents that might want to determine an optimal strategy for said game and execute on it.  (Here we have the understanding that an agent is an entity playing a game that possesses its own inherent agency - hence, "agent" - and has an objective in the context of the game with respect to some form of internally modelled utility function.)

Certainly, in the real world, agents will not have total information regarding the context of a particular game, and may not be rational.  Therefore, it makes sense to find a way to encourage agents in the aggregate to follow a particular strategy.  i.e., we would like our agents to be subjected to constraints which may be enforced to a greater or lesser extent.

\begin{dfn} (Action / Law)  A \emph{(system) action} or \emph{law} is a construct out of data relating to some form of summary statistics for a game over a population of agents.  Such an action can be used as a blunt tool to alter and/or constrain agent behaviour.  \end{dfn}

So in this sense we think of the \emph{system} within which the agents are playing, \emph{as itself a form of agent} - i.e., "system as agent" - and we are interested in determining two things: one, what policies and/or constraints such a system can employ to regulate behaviour of subsidiary agents within said system; and two, determination of a process to optimise when and how much to utilise such constraints.

\begin{dfn} (Policy) A \emph{policy} is an approach that a system agent takes to determine when and what actions to take towards particular subsidiary agents.  i.e. when and how to enforce laws.  The goal of a policy is to encourage subsidiary agents to behave in ways which are closer to an optimal strategy on the level of a particular set of coalitions of agents, i.e. at the level of concern of the system agent. \end{dfn}

In particular, hearkening back to towards the end of \cite{[Go5]}, one might be interested in looking at expressions such as $g_1 \circ g_0$, or $\circ ( g_2 \circ g_1 \circ g_0 ; h )$ (associated say to composition of information functionals into a more complex overarching information functional), and making these statements about groups or algebras in some logical sort of way, possibly as invariants of same.  One might intuit that this should be the logical framework to angle towards in terms of figuring out how to construct policies.

Naturally, a regulator should act in different ways according to different situations.  Types of situations arise in a natural emergent and combinatorial fashion according to the subtlety and nature of the game.  Interestingly, in the analysis to follow we find that to zeroth order we find three natural types of situations, and to first order we find there to be eleven.  Methods for formulation of optimal policy in regards to these different \emph{types} of situations will be sketched in this paper in terms of solution of appropriate natural PDE arising from criticality of certain information functionals.  (n.b. there is a nebulous link here to Type theory in programming, wherein to zeroth order we have String, Boolean and Float types, and to first order we have types associated to programming languages such as J or Haskell where we have function types. (The function types themselves would be $( \{ T \vert T : \{ String, Bool, Float \} \rightarrow \{ String, Bool, Float \}\}, Dict, Array)$, making eleven in total.))

One key goal of this paper will be to first more formally construct a description of a policy manifest for regulators / custodians / arbitrators / adjudicators in terms of an appropriate algebraic framework, and follow this with a description as to what situations are associated to regulation of situations concerning primarily 0-, 1-, and 2-agents (here a 0-agent makes decisions according to one statistical distribution (frequentist decisions), a 1-agent makes decision according to two (bayesian decisions), and a 2-agent makes decisions according to three).  The aim will then be to look into description of information theoretic techniques involving construction of information inequalities and setting their first variation to zero (a fairly standard technique) in order to describe how to compute optimal policy regarding intervention in various types of situations by a regulator.

\section{Overview}

In the above, it was suggested that we should look to algebraic considerations in order to abstract away unnecessary topological structure from \cite{[Go4]} and \cite{[Go5]}, and instead focus on the structural considerations associated to having two different layerings of agents - at player level and at system level.  This section will sketch roughly how we can construct such a framework, how this can be used to build policies to zeroth, first, and second order, and also very very briefly touch on more abstract concerns.  In particular, the aim of this section is to provide a rough compass for more detailed argument later within this document.

\subsection{Basics}

One can think of an optimal policy as a geodesic over the space of types associated to an algebra.

More precisely, we are interested in quantifying a choice of structure over ways to apply a type to a space.  Types can be things like $\circ$ and $\wedge$ with actions $\circ(f, g) := f \circ g$, and $\wedge(f, g) := f^g$, for instance.

We define a prototypical geometry on the space of types over a space $M$ as follows.  Let $\sigma$ be a metric over $M$, and $\tau$ be a metric over the space of metrics on $M$.  Then it makes sense that $\tau$ defines a geometry over the types associated to $\sigma$ by the following rough intuition: 

\begin{itemize}
\item[1)] we have multiple types available to us, and the simplest, and therefore most representative of the information of the system to first order (composition, exponentiation, addition and multiplication) form a 'type space' of dimension 4.
\item[2)] we can chain these types together so that we can select an arbitrary point in $\mathcal{Z}^4$.
\item[3)] we can broadly extend this to non-integers if we broaden our definition of type to include fractional applications of same; in this way we cover a 4-manifold of potential options.
\item[4)] this manifold admits a metric, $\tau$, that is dependent on the metric $\sigma$ in the original space that the type has been built on top of.  This metric $\tau$ must therefore be in the space of metrics of metrics over the original space.
\end{itemize}

This leaves us with this natural expression for a prototypical 'signal function' for such structures:

\begin{center} $f(m \vert a, b) := \delta(\delta(\tau \circ \sigma(m) - b) - a)$ \end{center}

and, as mentioned above, this defines a geometry over the space of types.  Defining the notation $\bardelta_{a}(\sigma(m)) := \delta(\sigma(m) - a)$, we can rewrite the above as

\begin{center} $f(m \vert a, b) := \bardelta_a \bardelta_b ( \tau \circ \sigma(m) )$ \end{center}

Now, to pare away unnecessary structure, and push our construction more towards that of an algebraic bent: let now $m$ be in some Lie Group $M$.  Then $TM$ is a Lie Algebra, and $\tau$ defines an action from the product of the Lie Algebra with itself to the reals; so a 'metric' on the Lie Algebra $T_m M$ for each $m \in M$.  In this sense we have a basis for discussion of Algebraic Information Theory.

\subsection{Classification}

This gives us a good starting point, but it is by no means the end of the matter.  For it is natural within this form of abstraction to consider the classes of particular structures that can be built and/or extended from this basic structure.  Degree one extensions are easy; there are three.  In fact, it is a simpler than the above mentions of type theory like geometries, since we are interested in the partition of merely one nested Dirac delta, therefore:

\begin{itemize}
\item \begin{center} $f(m \vert a, b) := \delta(\sigma^{(1)}(m) - a))$ \end{center}
\item \begin{center} $f(m \vert a, b) := \delta(\sigma(m, n) - a))$ \end{center}
\item \begin{center} $f(m \vert a, b) := \delta(\sigma(\phi) - a))$ \end{center}
\end{itemize}

where $\sigma^{(1)}$ is a function from the space of metrics to metrics, and $\phi$ is a function from $M$ back to itself.  I claim that these are representative of potential system actions relevant to the consideration of regulating behaviour of 0-agents.

Things get more interesting when we are interested in extensions of degree 2.  When we endeavour to classify the resultant topologies generated, with an emphasis to focusing on natural generalisations / extensions (more to come on this later, when we look at Ramsay theory), we determine eleven potential topologies, as demonstrated by \ref{tab:1}, bearing in mind that the Fibonacci sequence is a useful first order sequence to use in order to inductively construct structure - therein there being a natural build of 2 extra structures at ground, 1 extra structure at level 1, and 1 extra structure at level 2.

%\begin{itemize}
%\item $\sigma^{(1)}(\tau, \phi)(m)$
%\item $\sigma(\tau, \phi^{(1)})(m)$
%\item $\sigma(\tau, \phi)(m^{(1)})$
%\item $\sigma(\tau, \phi)(m, n)$
%\item $\sigma^{(1)}\tau(m, n)$
%\item $\sigma \tau(m^{(2)})$
%\item $\sigma \tau^{(1)}(m^{(1)})$
%\item $\sigma^{(1)}\tau(m^{(1)})$
%\item $\sigma \tau(m, n^{(1)})$
%\item $\sigma \tau^{(1)}(m, n)$
%\item $\sigma \tau(m, n, p)$
%\end{itemize}

\begin{table*}[t]
  \centering
  \begin{tabular}{lll}
    1. $\sigma^{(1)}(\tau, \phi)(m)$ & 
    5. $\sigma^{(1)}\tau(m, n)$ & 
    9. $\sigma \tau(m, n^{(1)})$ \\
    
    2. $\sigma(\tau, \phi^{(1)})(m)$ & 
    6. $\sigma \tau(m^{(2)})$ & 
    10. $\sigma \tau^{(1)}(m, n)$ \\
    
    3. $\sigma(\tau, \phi)(m^{(1)})$ & 
    7. $\sigma \tau^{(1)}(m^{(1)})$ & 
    11. $\sigma \tau(m, n, p)$ \\
    
    4. $\sigma(\tau, \phi)(m, n)$ & 
    8. $\sigma^{(1)}\tau(m^{(1)})$ 
  \end{tabular}
  \caption{Allowed topologies for partition \{2\} at depth 1}
  \label{tab:1}
\end{table*}

This is not the complete picture actually, for it is natural to consider the set of all partitions of 2: $\{ \{2\}, \{1, 1\} \}$, and thereby consider structures of the following kinds:

\begin{itemize}
\item $f(m \vert a, b) := \bardelta_a \bardelta_b ( \tau \circ \sigma(m) )$
\item $f(m, n \vert a, b) := \bardelta_a \sigma(m) \bardelta_b \tau(n)$
\end{itemize}

where as above there are 11 potential applicable topologies for the first.  For the second, there is 1 potential applicable topology.  Why one, and not three?  Because since we are considering higher abstraction (interaction with the metagame), we need to pare away some complexity.  In particular, it makes sense to pare away $p(1) = 2$ weights for the second class of structures, leaving $p(1 - p(1)) = p(-1) = 0$ weights to distribute, and therefore a total combinatorial complexity of 1.  This makes for a total description at this level of abstraction of size 12.

\subsection{Interpretation}

In game theory, we will say that a policy or law is a \emph{constraint} on possible moves in a game, usually with the intent on serving the interests of a group to whom all players belong.

The number of degrees of freedom, or situational topologies, are scenarios or modes of action by players in the game (simplifying from coalitions of same, since we are focusing on the algebra here, rather than CW structures for players).

Construction of optimal policies or a set/code of laws must needs consider each of these topologies individually.

An optimal policy should also take into account externalities or sources of \emph{excession}, which may perturb the course of the game.  To clarify,

\begin{dfn} (Excession) \emph{Excession} is the role of agency beyond the scope of a game by unknown players, which can nonetheless exert substantial influence on the way it is played, or even potentially completely change the rules and/or scope of play at certain turning points, after the manner of same in, say, catastrophe theory.  In particular, players of a game experience an \emph{excession event} if/when such an action of external agency becomes more apparent.  \end{dfn}

Therefore, we have the interpretation that the eleven potential topologies above correspond to a first order classification of ways to restrict modes of action by players in a game.  We expect naturally to be able to construct information invariants for same, and thereby to intuit optimal policies or laws that optimally constrain the action of players in terms of serving the interests of the group, while also giving them (the players) incentive to choose to join / remain with the group with said constraints.

The one topology for the partition $\{1, 1\}$ of 2 corresponds to a classification of ways to respond to \emph{excession} in a way that optimally guarantees the interests of the group.

On a higher order level, we might of course have law to law interactions.  If we are interested in more sophisticated concerns, in order to understand guiding principles for adjudication of a game in the context of some suitably defined metagame, we do need to bear such in mind (we will in fact find that the 11 potential topologies above becomes 42 in said instance).  

In particular, we have for particular partitions of 3, being $\{3\}, \{2, 1\}, \{1, 1, 1\}$ that:

For $\{1,1,1\}$ there corresponds structures $f(m, n, p \vert a, b, c) := \bardelta_a \sigma(m) \bardelta_b \tau(n) \bardelta_c \phi(p)$, with potential topologies of multiplicity 1.

For $\{2,1\}$ there corresponds structures $f(m, n \vert a, b, c) := \bardelta_a \bardelta_b ( \tau \circ \sigma(m) )\bardelta_c \phi(n)$, with potential topologies analogous to the previous section of multiplicity 11.

For the singleton set $\{3\}$ there corresponds structures $f(m \vert a, b, c) := \bardelta_a \bardelta_b \bardelta_c ( \phi \circ \tau \circ \sigma(m) )$, with potential topologies of multiplicity 42.  These are covered in table \ref{tab:2}, bearing in mind a Fibonacci progression of 1, 1, 2, 3 at various levels therein.

\begin{table*}[t]
  \centering
  \begin{tabular}{lll}
    1. $\sigma^{(1)}\tau\phi(m_1, m_2, m_3)$ & 
    15. $\sigma^{(1)}(\tau_1, \tau_2)\phi m^{(1)}$ & 
    29. $\sigma\tau\phi(m_1, m_2^{(2)})$ \\
    
    2. $\sigma^{(1)}\tau\phi(m_1, m_2^{(1)})$ & 
    16. $\sigma^{(1)}\tau(\phi_1, \phi_2, \phi_3)m$ & 
    30. $\sigma\tau\phi^{(2)}m^{(1)}$ \\
    
    3. $\sigma^{(1)}\tau\phi m^{(2)}$ & 
    17. $\sigma^{(1)}\tau\phi^{(2)}m$ & 
    31. $\sigma(\tau_1, \tau_2)(\phi_1, \phi_2, \phi_3)m$ \\
    
    4. $\sigma^{(1)}\tau\phi^{(1)}(m_1, m_2)$ & 
    18. $\sigma\tau\phi(m_1, m_2, m_3, m_4)$ & 
    32. $\sigma(\tau_1, \tau_2)(\phi_1, \phi_2^{(1)})m$ \\
    
    5. $\sigma^{(1)}\tau\phi^{(1)}m^{(1)}$ &
    19. $\sigma\tau\phi m^{(3)}$ &
    33. $\sigma(\tau_1, \tau_2)\phi^{(2)}m$ \\
    
    6. $\sigma^{(1)}\tau(\phi_1, \phi_2)(m_1, m_2)$ &
    20. $\sigma\tau(\phi_1, \phi_2)m^{(2)}$ & 
    34. $\sigma\tau^{(1)}(\phi_1, \phi_2, \phi_3)m$ \\
    
    7. $\sigma^{(1)}\tau(\phi_1, \phi_2)m^{(1)}$ & 
    21. $\sigma\tau(\phi_1, \phi_2)(m_1, m_2, m_3)$ &
    35. $\sigma\tau^{(1)}(\phi_1, \phi_2^{(1)})m$ \\
    
    8. $\sigma^{(1)}\tau^{(1)}(\phi_1, \phi_2)m$ & 
    22. $\sigma\tau(\phi_1, \phi_2)(m_1, m_2^{(1)})$ &
    36. $\sigma\tau^{(1)}\phi^{(2)}m$ \\
    
    9. $\sigma^{(1)}\tau^{(1)}\phi^{(1)}m$ &
    23. $\sigma^{(1)}\tau(\phi_1, \phi_2^{(1)})m$ &
    37. $\sigma(\tau_1, \tau_2)\phi(m_1, m_2, m_3)$ \\
    
    10. $\sigma^{(1)}\tau^{(1)}\phi(m, n)$ & 
    24. $\sigma\tau(\phi_1, \phi_2^{(1)})(m_1, m_2)$ &
    38. $\sigma(\tau_1, \tau_2)\phi(m_1, m_2^{(1)})$ \\
    
    11. $\sigma^{(1)}\tau^{(1)}\phi m^{(1)}$ & 
    25. $\sigma\tau(\phi_1, \phi_2^{(1)})m^{(1)}$ &
    39. $\sigma(\tau_1, \tau_2)\phi m^{(2)}$ \\
    
    12. $\sigma^{(1)}(\tau_1, \tau_2)(\phi_1, \phi_2)m$ & 
    26. $\sigma\tau\phi(m_1, m_2, m_3^{(1)})$ &
    40. $\sigma\tau^{(1)}\phi(m_1, m_2, m_3)$ \\
    
    13. $\sigma^{(1)}(\tau_1, \tau_2)\phi^{(1)}m$ &
    27. $\sigma\tau\phi(m_1^{(1)}, m_2^{(2)})$ &
    41. $\sigma\tau^{(1)}\phi(m_1, m_2^{(1)})$ \\
    
    14. $\sigma^{(1)}(\tau_1, \tau_2)\phi(m_1, m_2)$ & 
    28. $\sigma\tau\phi^{(2)}(m_1, m_2)$ &
    42. $\sigma\tau^{(1)}\phi m^{(2)}$
  \end{tabular}
  \caption{Allowed topologies for partition \{3\} at depth 2}
  \label{tab:2}
\end{table*}

This makes for overall a description of total multiplicity 42 + 11 + 1, or 54 at depth / subtlety 2.

Here, 42 refers to optimal ways to regulate behaviour between players in said game, if these players are 2-agents.  11 refers to optimal ways for said players to exert custodial responsibilities over 1-agents, bearing in mind principles of excession.  1 refers to the optimal way to approach matters of excession to the domain of operability for 2-agents.

(From our previous discussion, we might ask why we only distribute $2$ weights at the second level, and none at the third.  This follows because we have a sequence $p(2), (p(0), \{ p(1), p(-1) \}), (\circ (p, p(1))(1), \{ p(-1), p(-1), p(-1) \})$, such that the sum of the components is the same in each part of the sequence.  The vague intuition here is that information about the metagame should be coarser for a faithful description, and information about the metametagame should be even vaguer.)

\subsection{What lies ahead}

Consider now the four tuple of manifolds $N := (M_{\mathcal{G}}, M_{\mathcal{A}}, M_{\mathcal{T}}, M_{\mathcal{\phi}})$.

Let $\phi, \psi \in \{ \epsilon \vert \epsilon : N \rightarrow N \} =: N^{(1)}$, and $K = N^{(1)} \times N^{(1)}$.  Furthermore, take $\sigma : TK \times TK \rightarrow R$.  Let $\tau$ be a metric over metrics over $K$ to $R$.

Then

\begin{center} $f(\phi, \psi \vert a, b) := \bardelta_a \bardelta_b (\tau \circ \sigma( \phi, \psi ) )$ \end{center}

is the key object of study.  This is the foundation of the study of torus categories, the next level of abstraction beyond what is covered in this paper.  We are also interested in zero sum Ramsay theory, which concerns a combinatorial approach towards examination of meta-mathematical concepts.  In particular, we expect to need to 'fold in' the idea of meta-mathematics into studies of structures (be they higher categorical / toposes, or otherwise) of higher order subtlety as an element of their mathematics.

\section{Policies for games}

In this section we will flesh out the theory of optimal policies for games concerning 0-agents.

\subsection{Initial formalism and intuition: A recap and a step forward}

Recall that in \cite{[Go4]} and \cite{[Go5]} we described a game formally as a system that had players who could make potential actions during each game 'turn'.   Indeed, within any game, for a given game turn, we should have potential actions or moves available to players.  The impact of these actions on the game is described by a decision metric $\sigma : T\Delta^3 \times T\Delta^3 \rightarrow R$, where $\Delta$ is a four tuple of CW complexes.  The information associated to this decision metric can be thought of as an expression for the payoff $g : \Delta^3 \rightarrow R$, and solving for the first variation of the information to zero (by the Cramer-Rao inequality) allows us to compute optimal strategies as geodesics of this system.  So far so good.

We eventually generalised this to a point where we had multiple information functionals in terms of theory spaces etc starting to be chained together - eg $g_1 \circ g_0$,  or $\circ ( g_2 \circ g_1 \circ g_0 ; h )$.  These expressions almost look like group multiplication.  So this lends us to ask the following question - is there a natural generalisation of this sort of structure to bring in ideas about (algebras of) types, etc?  Of course there is, but rather than thinking about metagames, 1-agents and so on from the get go, let us take a few further steps back and try to think about what we are trying to describe.

In particular, our previous concerns assumed that the game had a global set of actions and game rules which were available to players, but within this state of affairs, players (or coalitions of players, described by CW complexes) were free to do anything.  Now, let us drop the generalisation that players can be CW complexes, and consider them merely as points.  And let us introduce the notion that players can opt to be members of a subsystem that itself has agency to dictate and enforces constraints on the actions of such players.  Now, you might ask, why would a player willingly sacrifice agency to join the ranks of an organisation with such membership conditions?  In analogy with how cities and societies work, we might say that the reason would be protection and also access to greater resources.  And also in analogy with cities and societies, we might call these constraints laws, and their implementation and/or enforcement by said society / organisation a policy.  We might also expect optimal levels of enforcement to not to be totally iron-fisted - indeed, it might be mutually advantageous to the society and its members to have some slack in said constraint.  nb. also, in the real world, people tend to assume things are enforced, because the societies that couldn't in the past lost the advantage of having laws and systems, and went into decline and/or were superseded by societies that could.  So it is a reasonable assumption to make.

That is the preliminary intuition.  Consequent to this, we are interested in formulation of a theory of constraints of play in a game.

So say that we have a set of allowed actions described by $\sigma : TG \times TG \rightarrow R$ (assuming for simplicity now that $G$ is a four tuple of Lie Groups $G_i$, $i = 0, 1, 2, 3$).  Let $\tau$ be a metric on the space of metrics over $G$.  Suppose that $m = (m_1, m_2, m_3, m_4)$ is a point in $G$.  Then we have the notion of a payoff function given in terms of a natural information functional, constructed from this distribution functional:

\begin{center} $f(m \vert a, b) := \bardelta_a \bardelta_b ( \tau \circ \sigma(m) )$ \end{center}

This payoff function which we will call $P$ takes the form

\begin{center} $P(f) := \int_{m}\int_{a, b}f (\partial ln (f))^2dmdadb$ \end{center}

Solving for $\delta P = 0$ we extract a PDE which we can solve to determine the optimal policy for implementation / determination of the optimal amount of 'slack' concerning a particular topological restraint associated to an understanding notion of law, given by the implicit choice of topology in construction of $P$.

Indeed, we have different choices of topology in construction of $P(f)$ which manifest in the construction of the statistical distribution itself.  To zeroth order the answers are different than to first order, which are given in \ref{tab:1}.  Indeed, to zeroth order, we have a very straightforward classification:

\begin{itemize}
\item 1. $f(m \vert a \vert (0, (0, 1))) = \bardelta_a ( \sigma(m, n) )$
\item 2. $f(m \vert a \vert (0, (1, 0))) = \bardelta_a ( \sigma(\phi) )$
\item 3. $f(m \vert a \vert (1, (0, 0))) = \bardelta_a ( \sigma^{(1)}(m) )$
\end{itemize}

So, where do the sequences $(0, (0, 1)), (0, (1, 0))$, and $(1, (0, 0))$ come from?  Largely speaking these come from a critical graph, or critical sequence - and the set of these sequences forms a set of 'trees', a concept which we could coin alternatively as a \emph{grove}.

\subsection{Sequences, subsequences, and critical graphs}

Broadly speaking, we are interested in a language that allows us to describe sequences of objects $S_{i}$ which themselves may either be numbers or in fact contain nested sequences $S_{(i, j)}$, which in turn, contain sequences $S_{(i,j,k)}$ ad infinitum.  And we are interested in \emph{critical sequences} in some way, shape or form.  Let us recast this another way - we can represent a sequence as a graph between $n$-simplices.  Leaving aside the matter of how graph multiplication is supposed to work for the time being, a graph admits a factoring if one can multiply two subgraphs together to get the graph back.

The rank of a graph is, vaguely speaking, the degree to which it deviates from being a chain of $0$-simplices.

Then there are three unique prime graphs such that their \emph{rank} is 1.  These can be represented as $(0, (0, 1)), (0, (1, 0))$, and $(1, (0, 0))$, and there are 11 prime graphs of rank 2.

There are 42 prime graphs of rank 3, as evidenced by \ref{tab:2}.

It is naturally of course interesting to ask how we might be able to represent such trees in a more natural way in terms of our choice of notation.  But there are actually two choices of tree we can use to build structure: a tree of functions, or a tree of Dirac delta functions.  I illustrate this as follows:

%%%% Insert omnigraffle diagrams here %%%%%

Consequently, it makes sense to consider notation $\bardelta_a (T)$ for $T$ a tree.  More generally, however, we might be interested in considering $\bardelta_{S}(T)$ (and indeed things like $\bardelta_U \bardelta_V(T(A, B))$, where $U, V, A, B$ are trees, and $T$ is something slightly more exotic).  I will explore matters concerning the latter in a later paper wherein I visit Ramsay theory more carefully.  However, for now, it is perhaps instructive to adopt the former notation to our problem of note.

In particular, we can rewrite the above in terms of $T \in \mathcal{T}$, the finite set of prime trees of rank 1, as:

\begin{center} $f(T \vert a, b) := \bardelta_a \bardelta_b (T)$ \end{center}

or more succinctly as

\begin{center} $f(T ; S) := \bardelta_{S}(T)$ \end{center}

where $S$ is the trivial tree with two nodes.

Let's try to flesh out this intuition a bit more.

First, we have the idea of prime trees.  Loosely speaking, we have a multiplication between trees.  A tree is prime if it cannot be factored into smaller non-trivial trees.  We also have a way to represent trees of structure for functions, and trees of structure for nested statistical distributions (which in our case are primarily Dirac delta functions).  The nodes of a tree correspond to different types of simplices, depending on the 'dimension' of the node, eg a 0-simplex, 1-simplex etc.

(A tree is prime if it cannot be factored into smaller trees.)

\begin{center}\scalebox{0.55}{\includegraphics{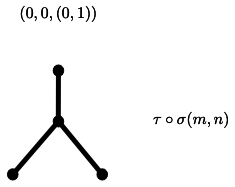}}\end{center}
\begin{center}\scalebox{0.55}{\includegraphics{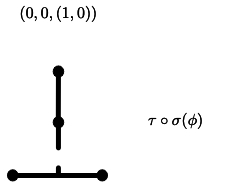}}\end{center}
\begin{center}\scalebox{0.55}{\includegraphics{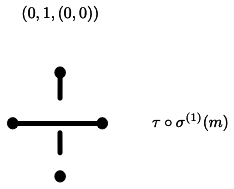}}\end{center}

Indeed the above is the most natural intuitive way to view of different allowable topologies for our algebraic information theories.  But we want to go further.

%\begin{center}\scalebox{0.25}{\includegraphics{../resources/tree/tree2.jpg}}\end{center}
\begin{center}\scalebox{0.75}{\includegraphics{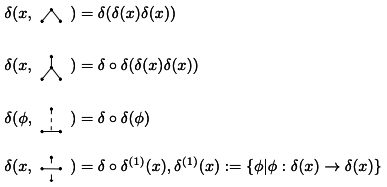}}\end{center}

Now $\delta_{T}(x) := \delta(x, T)$, with $T$ a tree can therefore be defined in general per the above interpretation.

Similarly, $\delta_a(T)$ can be defined with $T$ a tree on functions.

Consequently, more generally, $\delta_{S}(T)$ is well defined, for $S, T$ trees.  Therefore we have a 'tree of trees', where for each node in the tree of functionals, we have in general a tree of functions for its argument.
%\begin{center}\scalebox{0.25}{\includegraphics{../resources/tree/tree4.jpg}}\end{center}
%\begin{center}\scalebox{0.25}{\includegraphics{../resources/tree/tree5.jpg}}\end{center}

This leads us to the following important insight, expressing said ideas per the following notation:

Suppose that $\phi \in \{ \{ \kappa \rightarrow \kappa \} \rightarrow \kappa \}$, where $\kappa$ is the set of all trees.  So that $\phi$ is a map from the power set of trees - or the set of subtrees of a general tree - to a tree.  Then we can represent a treelike structure for Dirac deltas $S$ conjoined with a correspondence to the nodes of such trees with \emph{further} trees for functions, for each and every subtree of $S$ loosely as:

\begin{center} $\delta_{S, \phi}(args(\phi))$ \end{center}

where $args(\phi)$ is loosely the arguments of the aforementioned mapping as resolved at a particular subtree of $S$.  But this is unsatisfying notation.  Perhaps better and more suggestive is to overload the idea of a Dirac delta taking only one argument, and allow it to take two.

Then we have vaguely $\delta(S, \phi)$.  But then the second argument could \emph{itself} be a statistical distribution, so we could write

\begin{center} $\delta(S, \delta(\phi - a))$ \end{center}

which is highly suggestive.  Indeed, $S$ need not be a tree - it could be something else.  A fractional tree?  A point in a manifold?  A statistical distribution itself?

\begin{center} $\delta(\delta(\psi(m) - a), \delta(\phi(n) - b))$ \end{center}

%\begin{center}\scalebox{0.35}{\includegraphics{resources/tree/tree6.jpg}}\end{center}

But of course if we are allowing $\delta$ to take more than one argument, why not a countably infinite number of arguments?  So by dimension criticality assume that we have four arguments for $\delta$ that contain information.  And if this is to be a four dimensional space of four-manifolds, $(m_0, m_1, m_2, m_3) \in \bar{M}$, then we have an idea of some sort of matrix of metrics $\Lambda$, or a 4-tensor over them.

Then we are interested vaguely in structures of the form

\begin{center} $\delta( \delta_{ij}(\Lambda^{(ij)}(\bar{m}) - a_{ij}) )$ \end{center}

\subsection{The beginnings of higher category theoretic abstraction}

Ultimately, though, we are interested in computability.  So let's unwind things a bit, and recognise in fact that the information density measure

\begin{center} $\sigma(\Lambda)(m \vert a) := D_{ij} \partial_i ln(D_{kl}) \partial_j ln(D_{kl})$ \end{center}

where $D_{ij} := \delta(\Lambda^{ijk}(\bar{m}_k) - a_ij)$, has the property that this is a scalar, where $\Lambda$ is a 3-tensor.  So we can insert this neatly into a standard (non-overloaded) Dirac delta to have something computable:

\begin{center} $f(\bar{m} \vert a, b) := \delta(\sigma(\Lambda)(\bar{m} \vert a) - b)$ \end{center}

where $\bar{m} \in \bar{M} := (M_0, M_1, M_2, M_3)$ a vector of manifolds.  (The use of this information density furthermore allows us to have a form of generalised Cramer-Rao inequality holding true for the above statistical distribution - which would not be generically true if we did not use this particular form of nested structure.)

Finally, we might in fact be rather interested in $\phi \in \{ g : \bar{M} \rightarrow \bar{M} \}$, so that instead we would be studying

\begin{center} $f(\phi \vert a, b) := \delta(\sigma(\Lambda)(\phi \vert a) - b)$ \end{center}

But maybe it doesn't even make sense for $\sigma$ to be a 2-tensor at all, but rather a 3-tensor, in line with cybernetic considerations.  So this is a bit of an open question.  Nonetheless, the above can crudely be thought of as a starting point for the study of structures of deeper category theoretic complexity (in a sense, this seems logical, as said structure brings together and unifies depth two considerations with minimal geometry and topology (algebra) with maximal geometric structure, minimal depth and topology (geometry), and maximal topology, minimal geometry and depth (topology)).

Another open question is how to compute these things in a practical manner.  It is straightforward to construct algorithms for solution of 2-tensors, but 3-tensors and 4-tensors are wilier and harder to construct algorithms for that are computationally efficient.  This is of interest to us, because the above formulation actually is quite central to the consideration of construction of chat bots with adaptive trees.  

There may well be ways around this, such as through the use of moving from the idea of normal space to the space of special functions (like Beta functions, gamma functions, etc) and thereby compress the dimensionality of the above constructs into something more computable.  For instance, we might move to compress things by dealing with Meijer-G functions to move from 3 tensors to scalars, and from 4 tensors to 2 tensors - in a sense "lifting" to the space of functions viewed as analogous to surreal numbers with first uncountable cardinality, and identifying same with scalars through such a mapping.  Regardless, the detailed study of such structures as these shall be the subject of a later paper the author aspires to write on Ramsay theory, which will attempt to deal with the topic of combinatorial information theory in a way that hopefully would do it greater justice. 

For now, though, it is useful to understand roughly where our different topologies are coming from.  Later, in yet a different paper that shall concern itself regarding providing a map for exploration further, a sketch will be provided regarding how the ideas just mentioned herein fit in terms of the classification of structures of structure, as well as the classification of structures of structure of structure.  It will also be stressed as to the importance of lifting from the notion of primes being the atoms of structure to more abstract notions of irreducibility.  Such considerations may prove to be key in making progress with chat bots, as well as a number of other surprisingly practical and applied fields of endeavour.

\subsection{Determination of optimal policy}

Backtracking somewhat, recall that we are interested in distributions of the form

\begin{center} $f(m \vert a, b) := \bardelta_a \bardelta_b ( \tau \circ \sigma(m) )$ \end{center}

with an associated payoff function which we will call $P$:

\begin{center} $P(f) := \int_{m}\int_{a, b}f (\partial ln (f))^2dmdadb$ \end{center}

We have three different choices of topology in construction of $P(f)$ which manifest in the construction of the statistical distribution itself.  To zeroth order the answers are different than to first order, which are given in \ref{tab:1}.  Indeed, to zeroth order, we have a very straightforward classification:

\begin{itemize}
\item 1. $f(m \vert a \vert (0, (0, 1))) = \bardelta_a ( \sigma(m, n) )$
\item 2. $f(m \vert a \vert (0, (1, 0))) = \bardelta_a ( \sigma(\phi) )$
\item 3. $f(m \vert a \vert (1, (0, 0))) = \bardelta_a ( \sigma^{(1)}(m) )$
\end{itemize}

The sequences above are associated to a discrete variant of a deeper more general underlying subject (the study of combinatorial information theory, or Ramsay theory).  In order to determine optimal policy, we can solve each of these in turn and try to infer from a situation as to which is the most important in which circumstance.

However, more likely we would be interested in a constrained optimisation problem, where we attempt to optimise each and every one, subject to the others.

This brings us full circle back to Ramsay theory again.  Our distribution here is

\begin{center} $f(\phi \vert a, b) := \delta(\sigma(\Lambda)(\phi \vert a) - b)$ \end{center}

where we can think of the inner distribution as being akin to the distributions above via vague intuition.  Therefore, in order to determine the best policy to pull available policy levers associated to the relevant topologies, we would perhaps be better served if we were to instead try to optimise the information of this distribution.

Then we find that the geodesics through $\{ \bar{M} \rightarrow \bar{M} \}$ space (where $\bar{M} := M \times M \times M \times M$) associated to the geometry imposed by a $\Lambda$ and $\sigma$ that solve the PDE induced by $\delta I(f) = 0$ are paths through "policy lever pull space" that optimise choices (and levels of enforcement of) abstract policy used to constrain behaviour of agents within an abstract society within an abstract game, that optimise the competitiveness of said society.

It seems reasonable to posit that in the real world, societies and/or organisations that tend to do well are the ones that exhibit emergent behaviours, (often formed over significant spans of time) which reflect solutions of this rather abstract problem.

Note finally that we have only dealt with the zeroth order case.  The first order case and second order case are discussed briefly in the previous section (and tables provided), but the flow of the argument and the analysis remain roughly the same.

%%%%%%%%%%%%%%%%%%%%%%%%%%%%%%%%%%%%%%%%%%%%%%%%%%%%%%%%%%%%%%%%%%%%
%%%%%%%%%%%%%%%%%%%%%%%%%%%%%%%%%%%%%%%%%%%%%%%%%%%%%%%%%%%%%%%%%%%%
%%%%%%%%%%%%%%%%%%%%%%%%%%%%%%%%%%%%%%%%%%%%%%%%%%%%%%%%%%%%%%%%%%%%
%%%%%%%%%%%%%%%%%%%%%%%%%%%%%%%%%%%%%%%%%%%%%%%%%%%%%%%%%%%%%%%%%%%%
%%%%%%%%%%%%%%%%%%%%%%%%%%%%%%%%%%%%%%%%%%%%%%%%%%%%%%%%%%%%%%%%%%%%
%%%%%%%%%%%%%%%%%%%%%%%%%%%%%%%%%%%%%%%%%%%%%%%%%%%%%%%%%%%%%%%%%%%%
%%%%%%%%%%%%%%%%%%%%%%%%%%%%%%%%%%%%%%%%%%%%%%%%%%%%%%%%%%%%%%%%%%%%
%%%%%%%%%%%%%%%%%%%%%%%%%%%%%%%%%%%%%%%%%%%%%%%%%%%%%%%%%%%%%%%%%%%%

\raggedright
\bibliographystyle{plain}
\bibliography{../tex/references}

%%%%%%%%%%%%%%%%%%%%%%%%%%%%%%%%%%%%%%%%%%%%%%%%%%%%%%%%%%%%%%%%%%%%
%%%%%%%%%%%%%%%%%%%%%%%%%%%%%%%%%%%%%%%%%%%%%%%%%%%%%%%%%%%%%%%%%%%%
%%%%%%%%%%%%%%%%%%%%%%%%%%%%%%%%%%%%%%%%%%%%%%%%%%%%%%%%%%%%%%%%%%%%
%%%%%%%%%%%%%%%%%%%%%%%%%%%%%%%%%%%%%%%%%%%%%%%%%%%%%%%%%%%%%%%%%%%%
%%%%%%%%%%%%%%%%%%%%%%%%%%%%%%%%%%%%%%%%%%%%%%%%%%%%%%%%%%%%%%%%%%%%
%%%%%%%%%%%%%%%%%%%%%%%%%%%%%%%%%%%%%%%%%%%%%%%%%%%%%%%%%%%%%%%%%%%%
%%%%%%%%%%%%%%%%%%%%%%%%%%%%%%%%%%%%%%%%%%%%%%%%%%%%%%%%%%%%%%%%%%%%
%%%%%%%%%%%%%%%%%%%%%%%%%%%%%%%%%%%%%%%%%%%%%%%%%%%%%%%%%%%%%%%%%%%%

\appendix
\section{Why Algebraic information theory?}

Within this paper the choice has largely been to err somewhat towards provision of clear and crisp descriptions of particular mathematical structures for the sake of pedagogical clarity.  However, there is perhaps a risk that towards heading for polish in the above, that something might perhaps have been lost in an understanding as to underlying intuition.  Therefore, let's step back to the beginning once more and try to paint a different sort of perspective as to why one might think of what we are doing as 'algebraic information theory'.

\subsection{Foundations}

In order to conduct our analysis, we are interested in constructing various invariants for algebras and groups, and relating these to information invariants.  The prototypical example to consider is to consider maps $f : S^n \times S^m \rightarrow X$ (generalised simplicial homology), and look at the groups $C_{nm}$ consisting of homolopy equivalence classes of these.  There are boundary maps $\partial_{\rightarrow}$ that map from $C_{n+1, m} \rightarrow C_{n,m} \rightarrow C_{n-1,m}$ etc, and boundary maps $\partial_{\downarrow}$ that map from $C_{n, m+1} \rightarrow C_{n,m} \rightarrow C_{n, m-1}$ etc.  These form a grid of boundary operators linking groups together.  More generally, we can consider \emph{skip operators}, that map from $C_{n+1,m}$ to $C_{n-1,m}$ directly for instance.  It is slightly unclear to the author what the intuition there would be, though.

Regardless, the general idea is that with this construction we can construct an invariant 

\begin{center} $I(\mathcal{G}) := \Sigma_{\alpha, \beta, \gamma}g^{ij \alpha}\partial_{\downarrow}\partial_{\rightarrow}g_{ik \beta}\partial_{\downarrow}\partial_{\rightarrow}g_{jk \gamma} $ \end{center}

where $\mathcal{G}$ is the enveloping algebra consisting of all the homology groups $G_{n,m}$ defined as the quotient groups in such a way that a square is commutative, and $g_{nm \alpha}$ are the generators for $G_{n,m}$, and $g^{nm \alpha}$ relates to the cohomology generators dual to $g_{nm \alpha}$.

We can think of this roughly as an \emph{information}, and it satisfies a Cramer-Rao like inequality:

\begin{center} $I(\mathcal{G}) \geq 0$ \end{center}

Then $\delta I = 0$ allows us to determine an enveloping algebra $\mathcal{G}$ such that the information is critical.

But what does this have to do with policies?

\subsection{Application to game theory}

Consider as in \cite{[Go4]} a game $g : \Delta^{(0)} \times \Delta^{(0)} \times \Delta^{(0)} \rightarrow R$.  Recall that a point in $\Delta^{(0)}$ is a 4-vector $(p_0, p_1, p_2, p_3)$, where 

\begin{itemize}
\item $p_0$ is a representation of a coalition
\item $p_1$ is a representation of a decision for said coalition $a$ wrt a second coalition $b$
\item $p_2$ is the decision for coalition $a$ with a third coalition $c$
\item $p_3$ is the decision as to how coalition $a$ should account for competition between coalition $b$ and coalition $c$
\end{itemize}

and these are all CW complexes.

So far, so good.  A strategy is a choice of decision metric $\sigma : T\Delta^{(0)} \times T\Delta^{(0)} \rightarrow R$.  Suppose now that a rational strategy for $a$ is not rational with respect to an aggregate coalition of which $a$ is part.  Or alternatively that our coalition is not rational.  Then we need the idea of a policy.

Consider maps $f : S^{n} \times S^{m} \rightarrow \Delta^{(0)}$.  Then as above we can construct groups $G_{nm}$ as part of an enveloping algebra $\mathcal{G}$, with accompanying boundary operators.  The intuition for $f$ is that we are interested in extracting information regarding cliques amongst our players, and applying some form of forcing to counteract these cliques, which would otherwise make decisions that are suboptimal for the greater group ('society') of coalitions.

Since we are restricting degrees of freedom, instead of optimising a decision metric $\sigma$, we are interested in optimising the classes of a decision metric in relation to generators for our groups:

\begin{center} $[ \sigma ] : T\mathcal{G} \times T\mathcal{G} \rightarrow R$ \end{center} 

\subsection{Preliminaries}

Indeed, suppose $m$ is in some Lie Group $M$, with metric $\sigma$.  Since $M$ is a Lie Group, $TM$ is a Lie Algebra.  Let $TM$ be also itself a Lie Group and action a metric $\tau$.  In this sense we have a basis for discussion of Algebraic Information Theory.

Therefore, consider our fundamental structure for consideration of policies to be a group.  More than this, we would like it to be a Lie Group, so that group elements vary continuously over some space, ie, the group admits a differentiable structure with an atlas of charts.  This is motivated by say tetrated group multiplication under which $G$ should be closed, say $\star_{G} (a ; b)$, where $a$ is a group element and $b$ is a real number.  We also naturally are interested in a lower order multiplication between elements, $a \star_{G} b$ under which $G$ should also be closed.  That is:

\begin{itemize}
\item $G$ is closed under the operation of $\star_{G}$, and
\item $G$ is closed under iterated operation of $\star_{G}$ via $\star_{G} (g; r) \in G$, for all $g \in G$, $r \in R$.
\end{itemize}

For said group $G$, let $\mathcal{A}$ be its corresponding Lie Algebra.

We define a \emph{policy cost function} to be a map

\begin{center} $P : \mathcal{A} \times \mathcal{A} \rightarrow R$ \end{center}

eg, if elements in the algebra are generated by $x_1, \cdots, x_{k}$, then

\begin{center} $P : (x_{i_1}x_{i_2} \times \cdots \times x_{i_n}, x_{j_1}x_{j_2} \times \cdots \times x_{j_m}) \mapsto \mu \in R$ \end{center}

A density function over possible $P$ can be defined

\begin{center} $f(g, \alpha) := \delta(P(g) - \alpha)$ \end{center}

where $g \in G$, and $P(g) : \mathcal{A} \times \mathcal{A} \rightarrow R$ varies smoothly with $g$.

We can then define an information for $P$ in terms of our density function $f$ in a standard way:

\begin{center} $I(P) := \int_{G}\int_{\Omega}f(g, \alpha)(\partial ln ( f(g, \alpha)))^2 d\alpha dg$ \end{center}

This information will be non-negative.  Setting the first variation to zero then gives us an expression for $P$ which we can solve, in order to compute an optimal policy cost function $P_0$.

An optimal \emph{policy} is then a path in $G$ which follows geodesics of $P_0$.

This, however, is potentially naive.  Let us spell out a bit more intuition.

Say that $a := x_{i_1}x_{i_2} \times \cdots \times x_{i_n} \in \mathcal{A}$ is a policy.  Policies rather than agents are now considered to 'compete' with one another.  So $b := x_{j_1}x_{j_2} \times \cdots \times x_{j_m} \in \mathcal{A}$, $c := x_{k_1}x_{k_2} \times \cdots \times x_{k_{l}} \in \mathcal{A}$ are competing policies say.  Our task is to determine a way to determine the best strategy for a policy to take with its available 'levers' in order for it to have the maximal chance of success.

Then it seems more reasonable to define a policy cost function as a map

\begin{center} $P: \mathcal{A} \times \mathcal{A} \times \mathcal{A} \rightarrow R$ \end{center}

So what is a 'lever' available to a policy?  Surely this would be an action that it can take, if we consider a policy (as applied to a power set of CW complexes of agents in the overall pool of players) as an entity in its own right.  We wish moreover to ignore the underlying game and focus on the game of policy lever use optimisation.

But points in an algebra / in a group are trivial and have essentially no room for choice regarding levers, and are therefore not interesting.  Consequently it makes more sense perhaps to consider quotient groups $G/g$ of $G$ (as Lie groups) and their corresponding induced algebras $\mathcal{A}/\{g\}$.  Levers, then, are the available degrees of freedom or subgroups / induced subalgebras within the quotient group viewed as an entity.  Then we perhaps want to lift the concept of policy cost function to a map

\begin{center} $P: Pow(\mathcal{A}) \times Pow(\mathcal{A}) \times Pow(\mathcal{A}) \rightarrow R$ \end{center}

where $Pow(\mathcal{A})$ represents the set of all subalgebras of $\mathcal{A}$ including itself.  This is the definition that shall be used as the basis for the rest of the paper going forward.

So, what are the levers?  Orbits of the policy describe the levers/actions available to the policy.

\subsection{Iterated policy choices}

Suppose we have a situation where a policy wants to decide what levers to pull, subject to the understanding that in future it will have to do the same thing again many times over, and the information regarding what levers it has pulled will propagate into the decision trees of other competing policies.

Then the decision problem is different.  Solving the 'lever pull problem' for a policy in this iterated game is different to a 'one-off' choice.  We can follow the pattern of \cite{[Go4]} here and observe that possibly $5$ is the multiplicity that is important here.  However, $Pow(\mathcal{A})$ may not be sufficiently abstract.  It is possible indeed that consideration of the set $Pow^{(1)}(\mathcal{A}) := \{ f \rightarrow g \vert f, g \in Pow(\mathcal{A}) \}$ is a useful set to consider, and the expression $P: Pow^{(1)}(\mathcal{A}) \times Pow^{(1)}(\mathcal{A}) \times Pow^{(1)}(\mathcal{A}) \times Pow^{(1)}(\mathcal{A}) \times Pow^{(1)}(\mathcal{A}) \rightarrow R$ is the correct payoff function for this situation.

Actually there are three lifts to iteration:

\begin{itemize}

\item The first is as above.  Consider $Pow^{(1)}(\mathcal{A}) := \{ f \vert f: Pow(\mathcal{A}) \rightarrow Pow(\mathcal{A}) \}$, functions from the power set of $\mathcal{A}$ to itself.
\item The second concerns $Pow(\mathcal{A}^{(1)})$, where $\mathcal{A}^{(1)}$ is the set of maps from $\mathcal{A}$ to itself.
\item The third consists of consideration of $Pow Pow \mathcal{A}$, or $\circ (Pow; p(1)) (\mathcal{A})$, where $p: N \rightarrow N$ is the prime number sequence.

\end{itemize}

Intuition for these three types is as follows:

\begin{itemize}

\item For the first type, this concerns games where the levers available to a policy are subject to the iterate of the game and whatever previous lever pulls were made by constituents.

\item For the second type, this concerns games where we have certain policies and we wish to make several lever pulls (iterated choice).

\item For the third type, this concerns iteration of selection of policies, wherein each policy can only make one move, but can seed offspring policies from its choice (parent-child sequence / hereditary selection).

\end{itemize}

This emergence of three different strands of structure in terms of determination of appropriate strategies for pulling policy levers relates to the classification of \emph{situations} within which policies are used.  A \emph{situation} is an arena wherein multiple policies are used to make decisions.

To clarify this notion: 

\begin{dfn} (Situations, arenas, and modes and resonances).  A \emph{situation} is an arena wherein policies operate.  Such arenas have different natural choices of structure that policies can be applied to - and a finite number of said choices.  These correspond to \emph{modes} or \emph{resonances} of the arena, which act as wrappers around a policy.  Said modes (interchangeably, resonances) correspond to a choice of modelling the action of a policy in said situation.  Every policy can operate in all natural modes in a situation, therein it is necessary to consider all relevant meta-structures (resonances) that potentially apply as wrappers around sets of such policies.

For iterated policy choices for a policy associated to a Lie Algebra $\mathcal{A}$, there are three:  $Pow^{(1)}(\mathcal{A}), Pow(\mathcal{A}^{(1)}), PowPow\mathcal{A}$.
\end{dfn}

We can think of situations equivalently as instances of a universal class $\mathcal{C}$ which maps from the set of types to a function space.  Above, we have three types $J$ that map from a Lie Algebra to a function space.  We are interested in characterisation of $\mathcal{C}$ in a natural way.  Let $\mathcal{T}$ be the space of types (in this case discrete).  Then we would like to understand maps $\mathcal{C}: \mathcal{T} \times G \rightarrow \{ f \vert f : \mathcal{T} \times G \rightarrow G \}$, on some natural Lie Group $G$.

In particular we would like to compute for instance $\mathcal{C}(J, g) \in G$.

Maybe let's forget Lie Algebras and Lie Groups, and just state things in terms of tangent spaces.  Let's have a function $f: M \rightarrow M$.  Now, let's \emph{mutate} it with respect to a type $J$ that acts on the tangent space $TM$.  For instance, if we mutate it with respect to a function $g$ in type $T$ corresponding to $Pow^{(1)}(TM)$, ie $g \in \{ g : TM \rightarrow TM \}$, we transform $f$ into (in this case) $f^g$, or $\wedge(f ; g)$, where $\wedge$ is a representation of this type.

More generally, the mutation of $f$ with respect to an instance $g$ of $J$ is some function $J(f ; g)$.

Consequently, the types corresponding to mappings for first order:
\begin{center} $Pow^{(1)}(\mathcal{A}), Pow(\mathcal{A}^{(1)}), PowPow\mathcal{A}$ \end{center}

correspond to 2-tuple pre-geometric operators $\wedge, \star$ and $\circ$ respectively, and we have that $\mathcal{C}(J): M \rightarrow \{ \{ M \rightarrow M \} \rightarrow \{ M \rightarrow M \} \}$ can be represented as 

\begin{center}
$\mathcal{C}(J, f) := J(f; )$
\end{center}

\subsection{More on Types}

We've considered prototypical types $(Id, \circ, \star, +)$.  But what about more complex types?

Certainly we can have types like $\mathcal{C}(g) := +(\circ(\circ(g; ); ); \star(g; ))$, but this might be too complicated.  Potentially we would instead like to consider types $\mathcal{C}(g, h)$, too, so that we can construct something like 'a metric' on type space, acting on the space of distributions dual to it.

So say we have two general types $\mathcal{C}, \mathcal{D}$ that are elements of the algebra generated by $(Id, \circ, \star, +)$ above.  Then we can recover numbers if we compute $\mathcal{C}$ with $\mathcal{D}^{T}$ as second argument.  eg if $\mathcal{C}(g) := \circ(g; )$, $\mathcal{D}(h) := +(h; )$, then $\mathcal{D}^{T} := +(; h)$ and $\mathcal{C}\mathcal{D}^{T}(g) := \circ(g; \sigma)+(\sigma; g)$ where $\sigma$ is a type metric.

So we move away from types to consideration instead of this nebulous 'type metric' $\sigma$ acting on two tuples of types, where $\sigma$ is a function of $g$, our statistical distribution over metrics in physical space; so that the underlying space for which $\sigma$ operates is a space of statistical distributions, and our types form a tangent space to this space of distributions.

We can thereby think of a statistical distribution for \emph{this} in turn, too:

\begin{center} $\mathcal{F}(g, \alpha) := \delta(\sigma(g) - \alpha)$ \end{center}

so that this becomes our way of thinking about geometry for algebra.

But we can think of this as a way to develop a handle on representing geometry for topology as well, if we think of types themselves as having geometry, with physical space as tangent, and the space of statistical distributions as the underlying space.

So in this way, we have a $\{$ rock, paper, scissors $\}$ representation for topology, geometry, and algebra, which is a natural way to think about unification of the ideas present in \cite{[Go1]}, \cite{[Go2]}, \cite{[Go3]}, \cite{[Go4]}, \cite{[Go5]}.

\subsection{Meta-Policies}

Extension of this idea leads one to examine situations for meta-policies (up to 11 structural variants) or meta-meta-policies (up to 31 structural variants).

For classification of situations concerning a meta-policy, we have:

\begin{itemize}

\item 0. $Pow(\mathcal{A}^{(1)} \otimes \mathcal{A}^{(1)})$

\item 1. $Pow^{(1)}(\mathcal{A} \otimes \mathcal{A}^{(1)})$
\item 2. $Pow^{(1)}(\mathcal{A}^{(1)} \otimes \mathcal{A})$
\item 3. $Pow Pow (\mathcal{A} \otimes \mathcal{A})$

\item 4. $Pow^{(2)}(\mathcal{A})$

\item 5. $Pow^{(1)}(\mathcal{A}^{(1)})$
\item 6. $Pow(\mathcal{A}^{(2)})$

\item 7. $Pow Pow^{(1)}(\mathcal{A})$
\item 8. $Pow^{(1)} Pow (\mathcal{A})$
\item 9. $Pow Pow (\mathcal{A}^{(1)})$
\item 10. $Pow Pow Pow(\mathcal{A})$

\end{itemize}

which makes for 11 situations.  An optimal strategy constructed to determine optimal play for a meta-policy within one of these 11 situations is a \emph{tenet}.  Together, these tenets form an \emph{ethical framework} or \emph{code of laws}.

Intuition for each strand is as follows: %fill in details here

\begin{itemize}

\item The 0th situation, $Pow(\mathcal{A}^{(1)} \otimes \mathcal{A}^{(1)})$: cooperation

The ethics as to how to cooperate or interact with other meta-policies.

This consists of subalgebras associated to $\mathcal{A}^{(1)} \otimes \mathcal{A}^{(1)}$.  Here we consider in essence $\mathcal{A}^{(1)}$ a superalgebra for policies acting on policies, or the space of meta-policies.  Therefore the object of interest to us are pairs of actions as a subgroup of the product superalgebra between two meta-policies.

This then indicates the optimal strategy for a meta-policy to interact with other meta-policies.

One could think of this as the approach for actions available to two-tuples of meta-policies, or interaction with other meta-policies.  In terms of more concrete intuition, this gives rules of engagement for dealing with alternative ethical frameworks, for instance.

\item The 1st situation, $Pow^{(1)}(\mathcal{A} \otimes \mathcal{A}^{(1)})$: response to action on meta-policy by a separate policy.

The ethics of response to action by a subentity.

\item The 2nd situation, $Pow^{(1)}(\mathcal{A} \otimes \mathcal{A}^{(1)})$: approach to action of meta-policy on a separate policy.

The ethics of action to produce response in a subentity.

\item The 3rd situation, $Pow Pow (\mathcal{A} \otimes \mathcal{A})$: approach to review of action of actions of a two-tuple of policies which are answerable to the meta-policy.

The ethics of review of the effect of the cascading effect of action due to actions between two subentities; how and when to legitimise action.

\item The 4th situation, $Pow^{(2)}(\mathcal{A})$: approach to approaches to approaches concerning action of a policy answerable to the meta-policy.

The ethics of how to construct an ethics concerning review of the actions of a single subentity.

\item The 5th situation, $Pow^{(1)}(\mathcal{A}^{(1)})$: approach to approaches regarding how to act.

The ethics as to how to act as a meta-entity, i.e., as to how the ethics should behave.

\item The 6th situation, $Pow(\mathcal{A}^{(2)})$: approach to externality/excession.

The ethics as to how to deal with super-entities which serve as excession relative to oneself.

\item The 7th situation, $Pow Pow^{(1)}(\mathcal{A})$: approach to action regarding approaches concerning the actions of an entity.

The ethics as to how to respond to plans concerning action formulated by an subentity.

\item The 8th situation, $Pow^{(1)} Pow (\mathcal{A})$

The ethics as to how to plan to respond to actions undertaken by an subentity.

\item The 9th situation, $Pow Pow (\mathcal{A}^{(1)})$

The ethics as to how to respond to the cascading events of one's actions.

\item The 10th situation, $Pow Pow Pow(\mathcal{A})$

The ethics as to how to respond to the cascading response caused by the cascading events due to the actions of a subentity.

\end{itemize}

\begin{rmk} A way of viewing these 11 tenets is as modes ('policies') for optimal play for 0-agents, given that same are subject to custodial stewardship by 1-agents, with a supervisory excession represented by 2-agents. \end{rmk}

Abstracting again, we have the idea of classes, and metaclasses, or inherited classes.

In particular, following the prior analysis, we are led to the idea of a map $\mathcal{C}$ that maps from metatypes to maps $\mathcal{D}$ that map from types to functions.  We are interested in 3-tuple operation at cybernetic level, and possibly 2-tuple operation at normal level.

\end{document}